\documentclass{article}
\usepackage{spconf,amsmath,graphicx}
\usepackage{amsfonts}
\usepackage{url}
\usepackage{array} % DOUBLE CHECK THIS PACAKGE is compatible
\usepackage{microtype} % DOUBLE CHECK THIS PACAKGE is compatible

\usepackage{enumitem}
\usepackage[table]{xcolor}
\usepackage{booktabs}
\setlist{nosep, leftmargin=14pt}
 
\title{Differentiable VQ-VAE's for Robust White Matter Streamline Encodings}
\name{Andrew Lizarraga$^{1}$\quad Brandon Taraku$^{2}$ \quad Edouardo Honig$^{1}$ \quad Ying Nian Wu$^{1}$ \quad Shantanu H. Joshi$^{2,3}$
}

 \address{$^{1}$ Department of Statistics and Data Science, UCLA, USA \\
     $^{2}$ Ahmanson-Lovelace Brain Mapping Center, Department of Neurology, UCLA, USA\\
     $^{3}$ Department of Bioengineering, UCLA, USA
}
\begin{document}
%\ninept
%
\maketitle
\begin{abstract}
    Given the complex geometry of white matter streamlines, Autoencoders have been proposed as a
    dimension-reduction tool to simplify the  
    analysis streamlines in a low-dimensional latent spaces.
    However, despite these recent successes, the 
    majority of encoder architectures only perform dimension reduction on 
    single streamlines as opposed to a full bundle of streamlines. This is a severe limitation of the encoder architecture that completely disregards the  
    global geometric structure of streamlines at the expense of individual fibers. Moreover, the latent space may not be well structured which leads to doubt 
    into their interpretability. In this paper we propose a novel 
    Differentiable Vector Quantized Variational Autoencoder, which are engineered to ingest entire bundles of streamlines as single data-point and  provides reliable 
    trustworthy encodings that can then be later used to analyze streamlines in the
    latent space. Comparisons with several state of the art 
    Autoencoders demonstrate superior performance in both encoding and synthesis.

\end{abstract}
\begin{keywords}
Streamlines, Diffusion Tractography, Differentiable, Gumbel Distribution, Vector Quantization
\end{keywords}
\section{Introduction}
\label{sec:intro}
Autoencoders (AEs), drawing inspiration from traditional factor analysis, have been successfully applied in data compression, segmentation, and representation tasks. However, their application in encoding high-dimensional structures, particularly white matter streamlines, has encountered emerging limitations \cite{streamline_ae_Zhong2022} \cite{streamnet_pmlr-v194-lizarraga22a}. While various dimension reduction techniques, such as UMAP and tSNE, have been explored for white matter streamlines \cite{chandio}, recent advancements in encoder architectures, notably Variational Autoencoders (VAEs) \cite{vae_DBLP:journals/corr/KingmaW13} \cite{feng2023variational} and Vector Quantized-VAEs (VQ-VAEs) \cite{vq_vae_NIPS2017_7a98af17}, have shown superior results in dimension reduction tasks. In light of these developments, our research focuses on leveraging these architectures to analyze white matter streamlines, aiming to overcome the shortcomings associated with traditional AEs.

The application of VAEs and VQ-VAEs to streamlines is not without challenges. VAEs, which strive to create meaningful latent encodings, encounter difficulties in optimization since encodings are required to be Gaussian. This process involves minimizing the KL-divergence through the Evidence Lower Bound (ELBO), a task that can result in noisy reconstructions. VQ-VAEs, on the other hand, eliminate the need to optimize the ELBO by using a uniformly distributed codebook of quantized vectors. The distribution for selecting codebook vectors is determined via an arg-minimization problem, making the KL divergence between codebook and selection distribution constant. Despite this advantage, VQ-VAEs introduce the issue of non-differentiable selection of codebook vectors, necessitating a straight-through estimator \cite{vq_vae_NIPS2017_7a98af17}. This means that the neural network can't backpropogate gradients to adjust the codebook vectors during training.

This has prompted techniques such as using an exponential
moving average (VQ-EMA) to adjust and improve utilization of the codebook vectors.
However, even with these additional improvements, the reconstruction results can still be noisy. 
%In order to 
Other proposals to
effectively sample the codebook vectors (post-trainnig) to provide high quality image reconstructions,
requires swapping the uniform prior on the codebook with a
%after training, typically one would then swap the uniform prior on the 
%codebook vectors with 
strong prior discovered by another model like PixelNet \cite{vq_vae_NIPS2017_7a98af17} 
or a Transformer \cite{vq_gan_Esser2020TamingTF}. But we 
don’t have such luxuries for white matter streamline analysis given that there are so few architectures trained on streamlines and the datasets are typically too
small to learn powerful auto-regressive models. To address these issues, we propose a
novel Differentiable VQ-VAE (VQ-Diff) which allows for a fully differentiable
approach to the quantization step in the traditional VQ-VAE. We demonstrate state
of the art results for streamline reconstruction and empirically observe the models
robustness to perturbations in the latent space suggesting that geometrically
similar streamlines are grouped in similar neighborhoods.

\subsection{Contributions}
This paper makes the following contributions:
\begin{itemize}
    
    \item We propose a novel neural network architecture (VQ-Diff) that improves upon the VQ/VAE models by enabling a differentiable approach.

    \item Our model avoids the need for optimizing the KL divergence as is done in traditional VAE based models.
    
    \item Our model parallels the reconstructive performance with AEs, yet yields a robust and reliable latent encodings of streamlines.
    
    \item Our model demonstrates superior reconstruction performance compared to the 
    state of the art VAEs, VQ-VAEs, and VQ-EMAs.

    \item We develop and release an open-source PyTorch dataset, derived from the {\em Tractoinferno} \cite{tractoinferno} dataset, but offering full latent space encodings based on our model VQ-Diff as well as other competing state of the art models.
\end{itemize}

\section{Methodology}
\label{sec:methodology}
Unlike 2D images, a bundle of white matter streamlines is a collection
of curves
$B = \{f_i : \mathbb{R} \rightarrow \mathbb{R}^3 | i = 1, \dots, N\}$. As a result,
streamline bundles exhibit heterogeneous patterns, making their representation using codebook vectors challenging. Our approach involves composing weighted combinations of codebook vectors $\{e_1, \dots, e_k\}$ , allowing for more flexibility in the VQ-models. Typically, a 
VQ-model assumes a uniform prior $p(x)$ on the codebook and sets 
$q(x)$ to be a codebook selection distribution that arises from solving 
$\text{argmin}_i ||z - e_i||_2^2$. This is done in VQ-models because the KL divergence 
between $p(x)$ and $q(x)$ will be constant \cite{vq_vae_NIPS2017_7a98af17} and therefore these
models don't need to optimize the ELBO which in turn produces less noisy reconstructions.
However for our VQ-Diff model we let $p(x) = \frac{1}{\sqrt{2\pi\sigma^2}}\exp(-\frac{x^2}{2\sigma^2})$ and $q(x) = \frac{1}{\beta}\exp(-\frac{x}{\beta} - \exp(-\frac{x}{\beta}))$
be a zero-mean Gaussian and Gumbel distribution respectively. 
Then we take the weighted combination of the codebook vectors given by $s_j = \sum_{i=1}^k w_i e_i$ as our latent representation, where $e_i \sim p(x)$ and $w_i \sim q(x)$. 
Here, we show for the first time that the KL divergence between the Gaussian and Gumbel
Distribution is constant: First assume $p(x)$ and $q(x)$ are zero-mean Gaussian and Gumbel distributions, as described earlier. Then the KL divergence is computed as follows:
\begin{equation*}
    \begin{aligned}
    & D_{KL}(p||q) = E_{p(x)} \left[ \log \left( \frac{p(x)}{q(x)} \right) \right] \\
    %&D_{KL}(p||q)= E_{p(x)} \left[ \log p(x) - \log q(x) \right] \\
    &= E_{p(x)} \left[ \log p(x) - \log q(x) \right] \\
    & = E_{p(x)} \left[ -\frac{1}{2}\log(2\pi\sigma^2) + \log \beta -\frac{x^2}{2\sigma^2}  + \frac{x}{\beta} + e^{\frac{-x}{\beta}} \right] \\
    & = -\frac{1}{2}\log(2\pi\sigma^2) + \log \beta -\frac{1}{2} + E_{p(x)} \left[  e^{\frac{-x}{\beta}} \right] \\
    & = \text{const} + \int \frac{1}{\sqrt{2\pi\sigma^2}} \exp \left(\frac{-x}{\beta}\right) \exp \left(\frac{-x^2}{2\sigma^2}\right)dx\\
    & = \text{const} + \int \frac{1}{\sqrt{2\pi\sigma^2}} \exp \left(\frac{-1}{2\sigma^2}\left(x + \frac{\sigma^2}{\beta}\right)^2 - \frac{\sigma^2}{2 \beta^2}\right)dx\\
    & = \text{const} + \exp \left( \frac{\sigma^2}{2 \beta^2}\right)
    = \text{const}\\
    \end{aligned}
\end{equation*}

Similar to  the VQ-VAE architecture, this is indeed an advantage as we do not need to optimize the ELBO. Additionally, since the Gumbel weighted sum is differentiable we may backpropogate gradients to update the codebook. Moreover, we may choose a flat Gumbel distribution to ensure the network utilizes all the codebook vectors and avoids codebook collapse \cite{vq_vae_NIPS2017_7a98af17}. In summary, we improve over the VQ-VAE's weaknesses by passing 
gradients to update the codebook vectors and the flat Gumbel distribution ensures we utilize all of the codebook 
vectors.
%\subsection{Related Work}
%Related research on discrete-VAE's (dVAE) \cite{dvae_dalle_pmlr-v139-ramesh21a} proposes a similar solution with 
%$p(x)$ being a uniform
%and $q(x)$ to be a Gumbel distribution so that when you 
%anneal the Gumbel distribution, it will approximate a differentiable version of
%the argmin operator (Sec. \ref{sec:methodology:subsec:architecture}) and the model collapses to a VQ-VAE. The primary benefit is 
%that straight-through estimator is no longer needed since the codebook vector selection is differentiable. 
%However as discussed earlier, these hard-assignments of codebook
%vectors are not necessarily a good distribution for streamlines. Additionally, since 
%the KL-divergence between the 
%uniform distribution and the Gumbel distribution is not constant, one needs to re-introduce the ELBO which 
%introduces difficulties that the standard VAE encounters with the ELBO.

\subsection{Architecture}
\label{sec:methodology:subsec:architecture}
The VQ-Diff architecture is comprised of a ResNet encoder and decoder with the bottleneck being a Gumbel Soft-max assignment of 
weights. Our architecture is modeled after the VQ-VAE, but as mentioned earlier in Sec. \ref{sec:methodology},
the codebook vectors, $e_k$, are initialized with a Gaussian distribution, $p(x)$. Additionally instead of solving $\text{argmin}_i||z - e_i||_2^2$ to assign a codebook vector to the encoded input $z$, we apply Gumbel Soft-max \cite{gumbel_jang2017categorical}
across all distances $||z - e_i||_2^2$ to assign Gumbel weighted selection of
the codebook vectors: $s_j = \sum_{i=1}^n w_i e_i$.

\begin{figure}[htb]
\label{fig:vq_diff_architecture}
  \centering
  \centerline{\includegraphics[width=\linewidth]{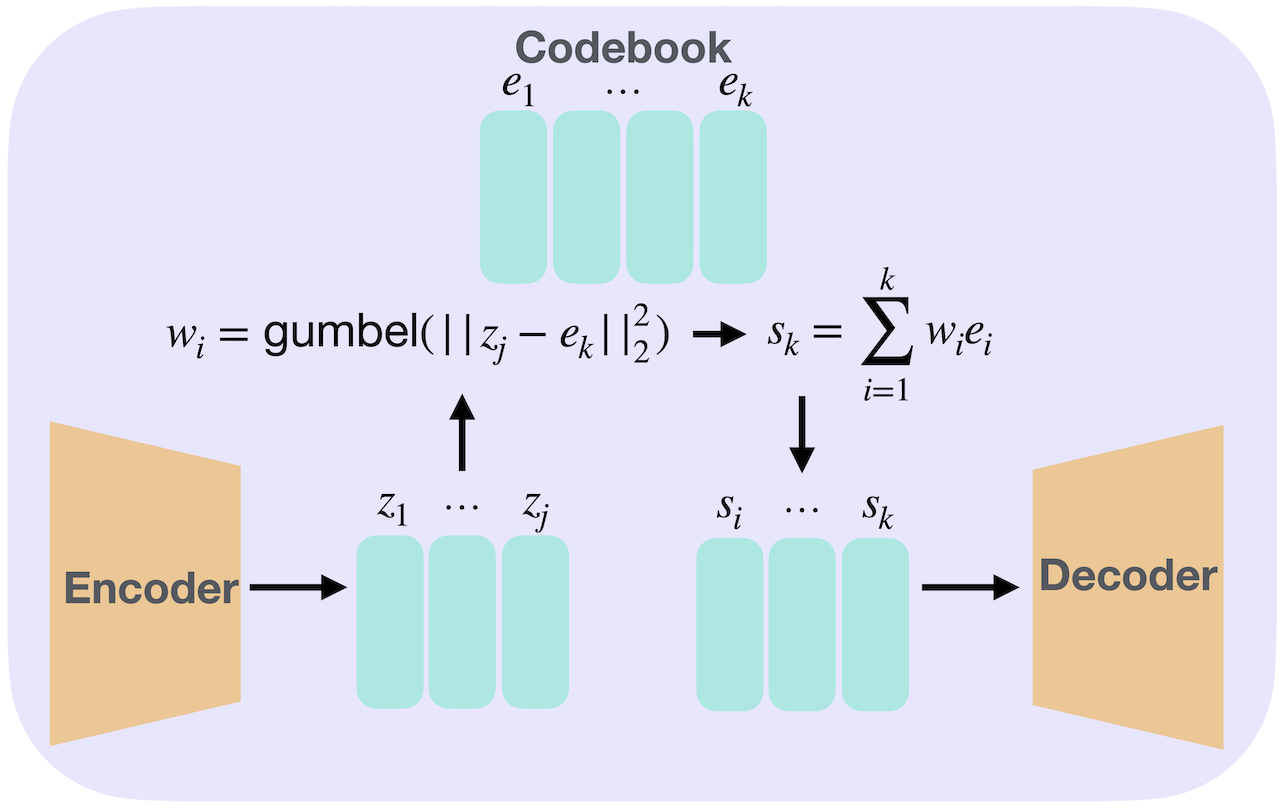}}
  \centerline{\textbf{Fig. \ref{fig:vq_diff_architecture}} Schematic of the VQ-Diff Architecture.}
\end{figure}

In summary, the encoder takes in a bundle $B$ and assigns a collection of latent vectors $z_j$ to each streamline. Then to 
each latent vector the
network assigns a gumbel weighted combination of codebook 
vectors: $z_j \mapsto s_j =\sum_{i=1}^n w_i e_i$, see Fig. \ref{fig:vq_diff_architecture}. The goal of 
the network is to learn a suitable codebook that captures 
features of streamlines that comprise a bundle.
We compare this architecture against an AE, VAE, VQ-VAE and a VQ-EMA all
composed of the same ResNet encoder and decoder.
The full model implementation can be found at  \url{https://github.com/drewrl3v/diff-vq-vae}.

\subsection{Data}
\label{sec:methodology:data}
We use the open-access dataset, {\em Tractoinferno} \cite{tractoinferno}, which
consists of 284 datasets acquired from a variety of 3T scanners, to demonstrate the performance of our model. Here, streamline
segmentation was performed with multiple techniques, resulting in 30 bundles per subject. In this paper, we only make use of the streamline coordinates as processed in {\em Tractoinferno} 
\cite{tractoinferno}. Since not all bundles comprise the same number of streamlines, we
selected tracts that consistently had over $1000$ streamlines which resulted in 12 bundles, namely, the Middle 
Cerebellar Peduncle (MCP), 
Right Frontopontine Tract (FPT\_R), 
Right Inferior Longitudinal Fasciculus (ILF\_R), 
Left Inferior Fronto-Occipital Fasciculus (IFOF\_L),
Left Frontopontine Tract (FPT\_L),
Left Inferior Longitudinal Fasciculus (ILF\_L),
Left Parieto-Occipital Pontine Tract (POPT\_L),
Right Inferior Fronto-Occipital Fasciculus (IFOF\_R),
Right Parieto-Occipital Pontine Tract (POPT\_R),
FrontalRostrum of Corpus Callosum (CC\_Fr\_1),
Left Pyramidal Tract (PYT\_L),
Right Pyramidal Tract (PYT\_R).

We then sub-divided each bundle per subject into groups of $64$ streamlines and up-sampled the number of points comprising a streamline 
to be $64$ points. Thus a single data-point for our neural network yields a $(64,3,64)$ tensor (has a dimension (number of streamlines $\times$ dimension (3) $\times$ number of points). 
This is done for computational convenience  to keep the  bundle size consistent and to allow the 
network to ingest $256$ batches of $64$ streamlines during training for a total of $16,384$ streamlines per training iteration. Since some tract produce more streamlines than others, we down-sample the number of bundles per tract to ensure there 
is an equal number of each bundle per tract. This prevents the network from favoring a particular bundle due to its overrepresentation in the training set. We then split the dataset into a $90\%$ train set, $10\%$ validation set. This PyTorch white matter streamline dataset is now open-access, and publicly available under the {\em Tractoinferno} \cite{tractoinferno} license at \url{https://github.com/drewrl3v/diff-vq-vae}. To the author's knowledge, this is the first such dataset that provides not only our full model and its parameters, but also the encoded streamlines and their latent spaces generated for state of the art models that have been used on streamlines. 

\subsection{Training}
\label{sec:methodology:training}
All models were trained for $15,000$ iterations each with a mean-squared error (MSE) loss function penalizing for low reconstructive quality of streamlines. Experimentally we found that a setting a Gumbel temperature of $\beta = 10.0$ and assuming the codebook prior 
to be Gaussian with variance $\sigma = 2.0$ produced the best results for the VQ-Diff model.
All models were ran on an AMD Ryzen Threadripper 3960X 24-Core Processor @ 3.8 GHz machine with a NVIDIA A6000 GPU and are released at: \url{https://github.com/drewrl3v/diff-vq-vae}.

\section{Experimental Results}
\label{sec:experimental}

\subsection{Reconstructive Quality}
\label{sec:experimental:subsection:reconstructive}
The Bundle analytic (BUAN) score \cite{buan_Chandio2020} is a state of the art method for comparing closeness of bundles of streamlines.
We used a very low threshold tolerance of $0.05$ for the bundle analytic score which makes the metric highly sensitive to minor 
differences among the bundles. 
After training we compared the BUAN scores across all the model architectures and bundles. 
A BUAN score closer to $1.0$ signifies perfect reconstruction. Table \ref{table:buan} is a record of the average BUAN score across all bundles in the validation set along with the first standard deviation in the BUAN score. As we can see, the VQ-Diff is on par with AE in terms of reconstructive quality, while VAE does not fare so well since the ELBO enforcing a Gaussian latent space is difficult to learn. 
As suggested in Sec. \ref{sec:methodology}, the VQ architectures, despite performing well in classical image reconstruction tasks  perform poorly on streamline data. 

This is because image intensities have a  neighborhood structure and may be assumed to be piecewise continuous with more relaxed geometric constraints,  while bundles of streamlines are composed of several individual fibers and have intrinsically complicated  geometry\cite{streamline_ae_Zhong2022}.

\begin{table}%[ht]
\label{table:buan}
\centering
\caption{BUAN Scores Across Architectures}
\resizebox{\columnwidth}{!}{%
\begin{tabular}{>{\bfseries}c >{\bfseries}c>{\bfseries}c>{\bfseries}c>{\bfseries}c>{\bfseries}c}
\toprule 
Bundle Name & VQ-Diff (Ours)      & AE                  & VAE                 & VQ-VAE              & VQ-EMA            \\
\midrule    
PYT\_R      & $0.9988 \pm 0.0038$ & $0.9999 \pm 0.0003$ & $0.4989 \pm 0.1923$ & $0.7154 \pm 0.1227$ & $0.6849 \pm 0.126696$ \\
PYT\_L      & $0.9988 \pm 0.0039$ & $0.9999 \pm 0.0007$ & $0.4764 \pm 0.2045$ & $0.7045 \pm 0.1179$ & $0.6778 \pm 0.119968$ \\
POPT\_R     & $0.9990 \pm 0.0034$ & $0.9999 \pm 0.0008$ & $0.4649 \pm 0.2166$ & $0.6867 \pm 0.1494$ & $0.6562 \pm 0.150048$ \\
POPT\_L     & $0.9989 \pm 0.0037$ & $0.9999 \pm 0.0009$ & $0.4815 \pm 0.2145$ & $0.6869 \pm 0.1397$ & $0.6508 \pm 0.144388$ \\
ILF\_R      & $0.9910 \pm 0.0114$ & $0.9999 \pm 0.0008$ & $0.2726 \pm 0.2019$ & $0.3409 \pm 0.1631$ & $0.2836 \pm 0.159151$ \\
ILF\_L      & $0.9910 \pm 0.0112$ & $0.9999 \pm 0.0007$ & $0.2399 \pm 0.2018$ & $0.3228 \pm 0.1561$ & $0.2408 \pm 0.149510$ \\
IFOF\_R     & $0.9888 \pm 0.0133$ & $0.9988 \pm 0.0012$ & $0.2808 \pm 0.2366$ & $0.3437 \pm 0.1970$ & $0.2793 \pm 0.188292$ \\
IFOF\_L     & $0.9855 \pm 0.0164$ & $0.9975 \pm 0.0016$ & $0.2340 \pm 0.2130$ & $0.3143 \pm 0.1773$ & $0.2296 \pm 0.162993$ \\
FPT\_R      & $0.9976 \pm 0.0058$ & $0.9999 \pm 0.0008$ & $0.2832 \pm 0.2102$ & $0.5063 \pm 0.1795$ & $0.4357 \pm 0.188509$ \\
FPT\_L      & $0.9971 \pm 0.0074$ & $0.9999 \pm 0.0006$ & $0.2689 \pm 0.2261$ & $0.4983 \pm 0.2025$ & $0.4335 \pm 0.209599$ \\
CC\_Fr\_1   & $0.9974 \pm 0.0110$ & $0.9999 \pm 0.0004$ & $0.3750 \pm 0.2033$ & $0.3380 \pm 0.1742$ & $0.2458 \pm 0.153786$ \\
MCP         & $0.9986 \pm 0.0043$ & $0.9999 \pm 0.0008$ & $0.3593 \pm 0.2528$ & $0.4989 \pm 0.2298$ & $0.3740 \pm 0.232267$ \\
\bottomrule
\hline
\end{tabular}
}
\end{table}
\begin{figure} %[htb]
\label{fig:orig_vs_recon}
  \centering
  \centerline{\includegraphics[width=0.98\columnwidth]{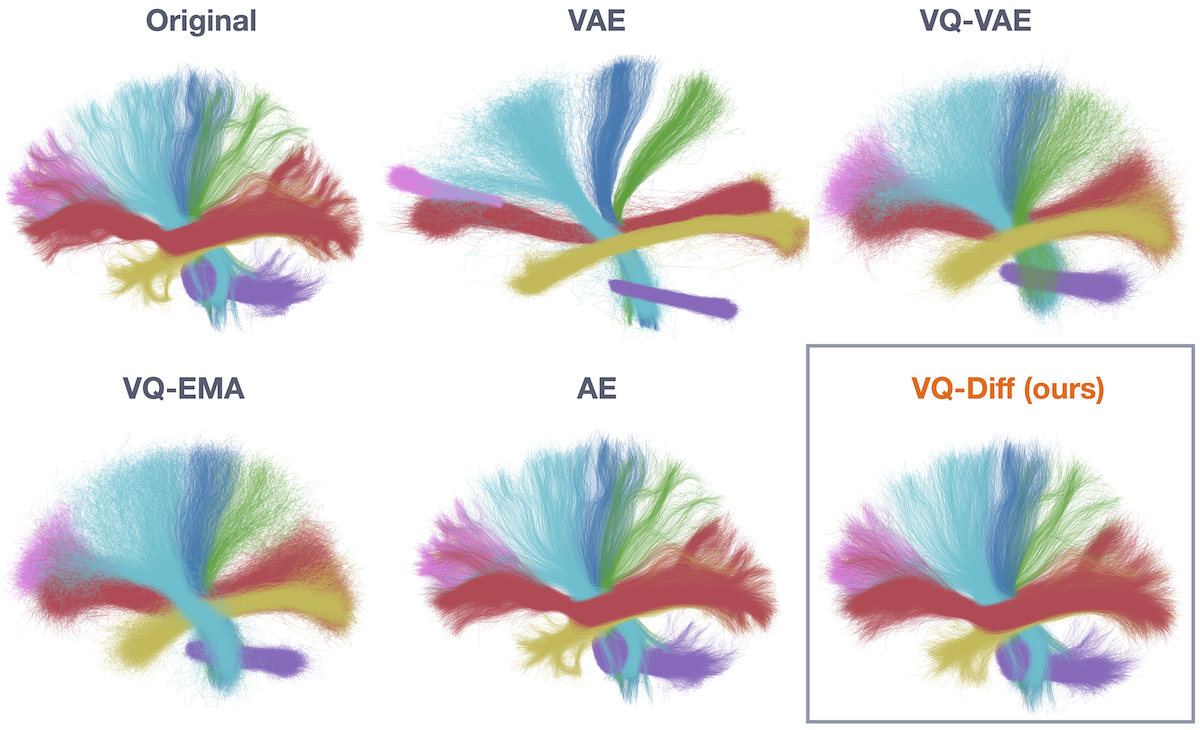}}
  \centerline{\textbf{Fig. \ref{fig:orig_vs_recon}} Full subject reconstructions across architectures.}
\end{figure}

%%%%%%%%%%%%%%%%%%%%%%%%%%%%%%%%%%%%%%%%%%%
\subsection{Visualizing the Latent Space}
\label{sec:experimental:subsection:latent_viz}
\begin{figure} %[htb]
\label{fig:latent_space}
  \centering
  \centerline{\includegraphics[width=\linewidth]{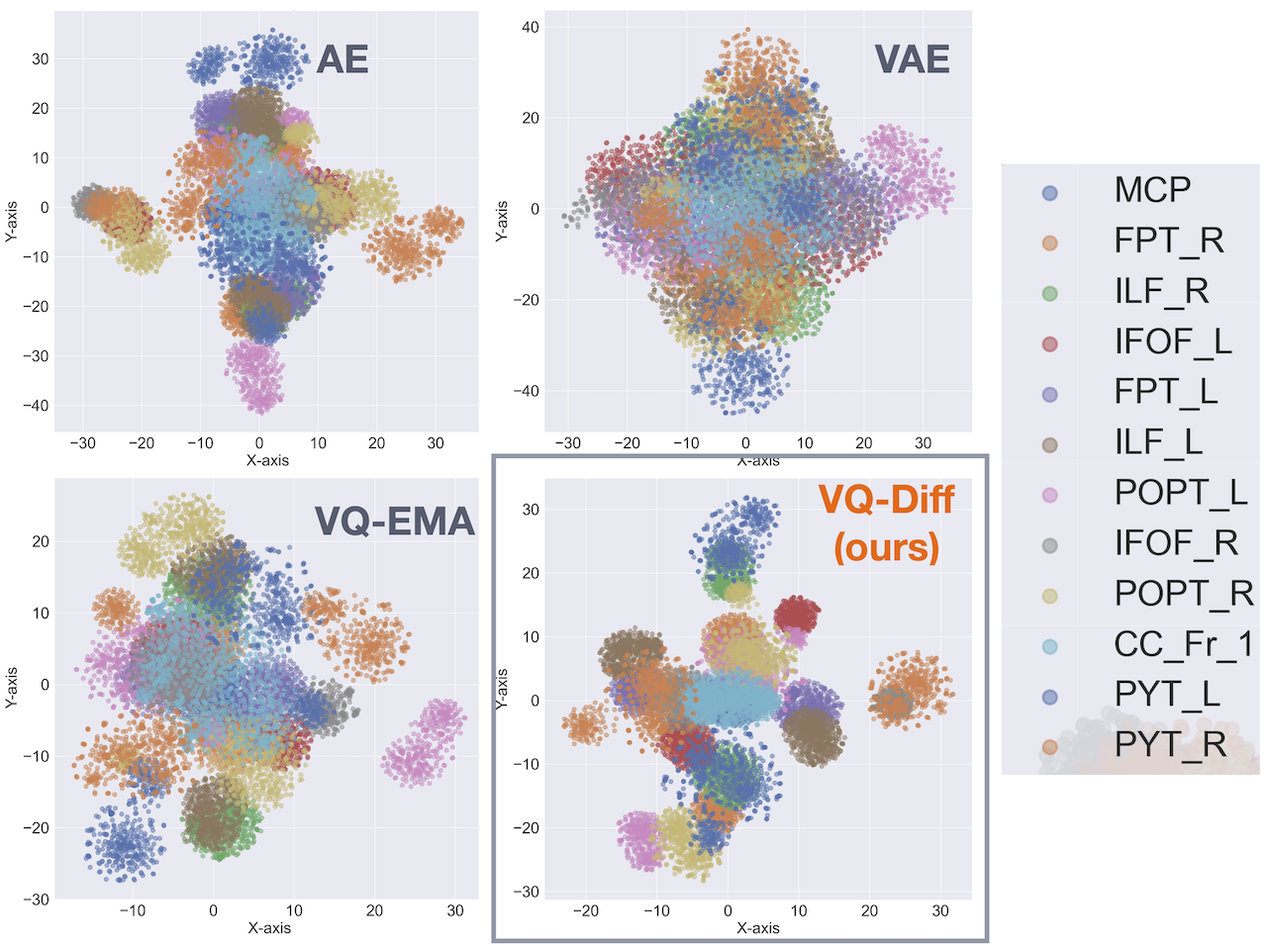}}
  \centerline{\textbf{Fig. \ref{sec:experimental:subsection:latent_viz}} Visualization of the latent space.}
\end{figure}

We visualize the latent spaces for each model by approximating the topology of the space by performing a tSNE \cite{tsne_JMLR:v9:vandermaaten08a} projection of the latent vectors, $z_j$. We see in Fig. \ref{sec:experimental:subsection:latent_viz} that the VQ-Diff is able to cluster respective bundle latent vectors and keep them roughly separated from other clusters. The VAE attempts to encode all latent vectors as Gaussian, which makes it difficult for the model to separate out categories, so we see a greater mixture of the latent vectors. The VQ-EMA is able to cluster the latent vectors but the clusters are more sparse. The AE on the 
other hand manages to cluster some of the latent vectors, but also mixes many of them in the center of 
the AE plot (Fig. \ref{sec:experimental:subsection:latent_viz}). It is noted that  
tSNE doesn't necessarily represent distances in the 
projected representation in Fig. \ref{sec:experimental:subsection:latent_viz}. To better 
understand the geometry, we instead isolate a latent vector and perturb it with noise. If the topology of the latent space is well regularized, then similar bundles should be mapped to a similar neighborhood. This means that reconstructions coming from a perturbed 
latent vector shouldn't drastically differ from the reconstruction coming from the original latent vector.
%The VQ-Diff plot in Fig. \ref{sec:experimental:subsection:latent_viz} suggest that it should tolerate pertubations given the tight clustering. We demonstrate that unlike the other encoder architechtures, VQ-diffs are remarkably resistant to pertubations in 
The VQ-Diff plot in Fig. \ref{sec:experimental:subsection:latent_viz}, which 
is tightly clustered for  bundles of the same type, but is able to achieve a separation across different bundle types, suggests  that it is robust to such perturbations. 
We perform an experiment to test this tolerance to perturbations in Sec. \ref{sec:experimental:subsection:perturbation}. 
% which reveal that, unlike other encoder architectures, the VQ-Diff exhibits remarkable resistance to these perturbations.

\subsection{Perturbation Analysis and Synthesis}
\label{sec:experimental:subsection:perturbation}
The reconstructive results of the VQ-Diff and AE are very promising. Given the strong reconstructive results for AE, Zhong et al.
\cite{streamline_ae_Zhong2022} and Legarreta et al. \cite{finta_LEGARRETA2021102126, gesta_LEGARRETA2023102761} have suggested that the latent space can be used to perform statistical analysis of streamlines. To explore the feasibility of these ideas, we perform perturbation analysis of the underlying encoded latent vectors
$z_j$. We choose the MCP bundle for demonstration purposes as it displays wide geometric variation in the population.

\begin{figure} %[htb]
\label{fig:perturb_mcp}
  \centering
  \centerline{\includegraphics[width=\linewidth]{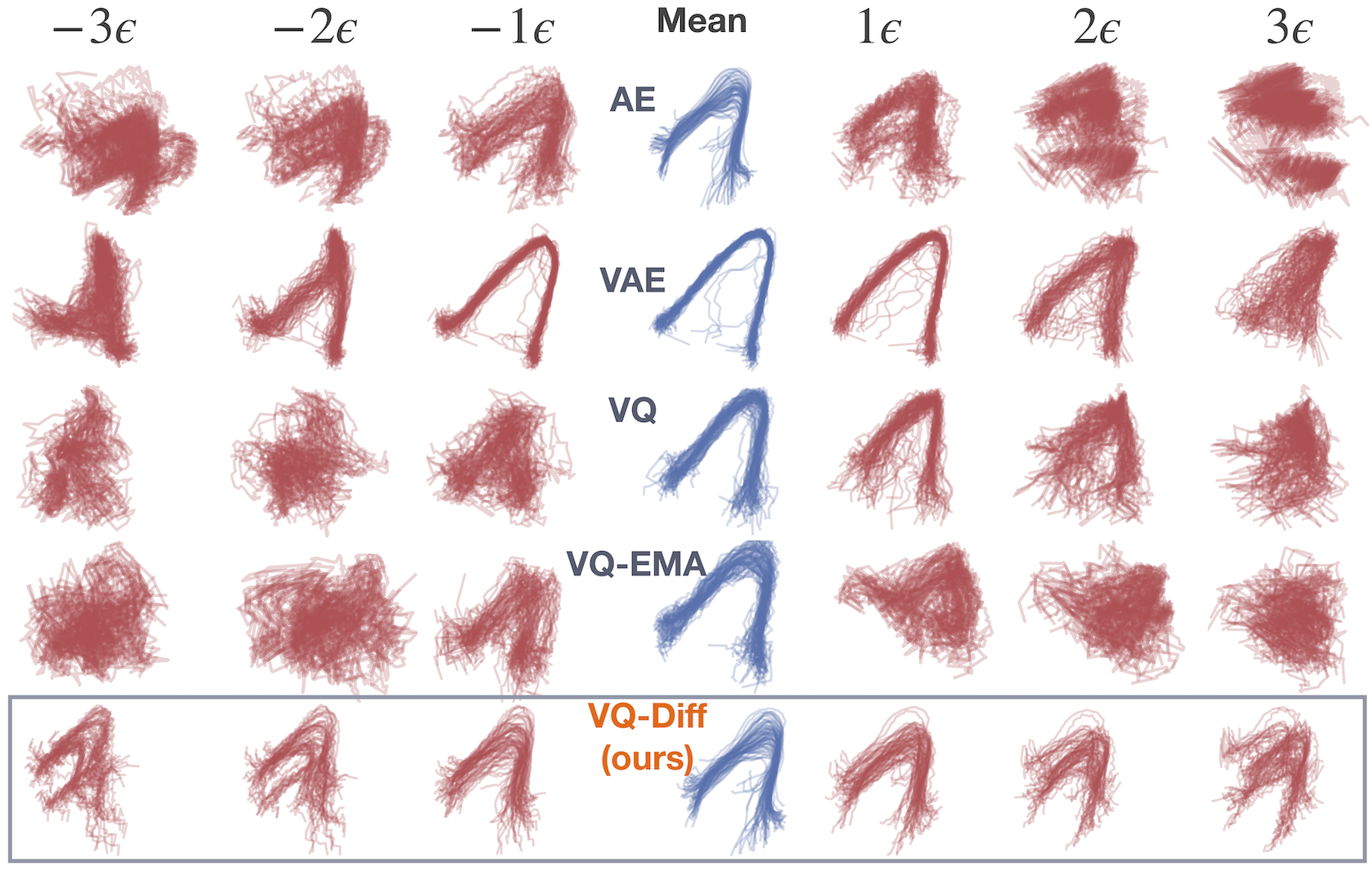}}
  \centerline{\textbf{Fig. \ref{fig:perturb_mcp}} Latent perturbations around the mean for the MCP bundle.} %\medskip
\end{figure}

Across all models we map the same MCP bundle for the same subject to their 
corresponding latent vector representations  $z_j$, then we perturb the vector 
by a small quantity: $z_j \mapsto z_j + \varepsilon$. We then 
pass $z_j + \varepsilon$ through each bottle neck layer 
for each architecture and reconstruct the bundle for each model architecture. As we see in Fig. \ref{fig:perturb_mcp}, given a selected latent vector for the MCP streamline 
bundle, the AE performs poorly when the latent vectors are perturbed by small noise, $\varepsilon = 0.5$, while the VAE model performs better than AE as expected. The VQ-VAE and VQ-EMA models behave better for extremely small perturbations at the mean, but quickly degrade in quality with increasing $\varepsilon$. The VQ-Diff model demonstrates superior tolerant to such perturbations across all models. This suggests that the geometry of the latent space for the VQ-Diff 
groups similar streamlines in the same neighborhood and in turn is a very robust latent representation that can be used for more  reliable distance analysis. 

\section{Discussion And Conclusion}
\label{sec:typestyle}

In this work we provide the following: A new open source PyTorch dataset derived from the 
{\em Tractoinferno} dataset, a novel Neural Network Architecture that has the state of the art
reconstructive performance of an AE, while also ensuring more robust latent representations. We demonstrate that the common assumption that the latent space of 
streamlines preserves local features does not hold for AEs. To the author's knowledge this is the first study to
analyze more recent AE architectures for
streamline analysis. 
We also observed that, while the reconstructive performance of the VAE is not on par with the AE or VQ-Diff, the 
Gaussian regularization on its latent space ensures that similar streamlines are within a neighborhood of the selected latent
vector for MCP. 

Overall, the VQ-Diff stands out as a highly robust architecture, having the potential to be trained across diverse MR image modalities. This flexibility underlines its potentially substantial impact in the field of medical imaging. In contrast, while the VQ-VAE and VQ-EMA exhibit limitations in effectively capturing the variability of streamlines in their codebooks, leading to lower reconstructive quality, they do offer a slightly more robust approach in terms of latent representations compared to the AE. This distinction highlights the unique strengths and weaknesses of these architectures, underscoring the VQ-Diff’s strengths as a particularly valuable tool  in medical imaging applications. 

%The VQ-VAE and the VQ-EMA are not able to effectively capture the variability of streamlines in their codebooks, 
%so their reconstructive quality is low, but the latent representations are slightly more robust than the AE. 
%
%The VQ-Diff is a very general
%architecture and can be trained on a variety of MR image modalities 
%making it's potential impact in medical imaging 
%very significant. 

\section{Compliance with ethical standards}
\label{sec:ethics}

This research study was conducted using  human subject data collected retrospectively and 
made available as an open-source project, {\em Tractoinferno} \cite{tractoinferno} \url{https://openneuro.org/datasets/ds003900/versions/1.1.1/download}. Thus ethical approval was not required under the licence attached to the open-source dataset.

\section{Acknowledgments}
\label{sec:acknowledgments}
This research was supported by the NIH NIAAA (National Institute on Alcohol Abuse and Alcoholism) awards R01-AA025653 and R01-AA026834 (SHJ) and was partially supported by the NSF DMS-2015577 award (YNW). 

\bibliographystyle{IEEEbib}
\bibliography{refs}

\end{document}